\documentclass[letterpaper, 10 pt, conference]{ieeeconf}  

\IEEEoverridecommandlockouts                              

\usepackage{cite}
\usepackage{amsmath,amssymb,amsfonts}
\usepackage{algorithmic}
\usepackage{graphicx}
\usepackage{booktabs}
\usepackage{textcomp}
\usepackage{xcolor}
\usepackage{soul}
\usepackage{orcidlink}
\usepackage{subcaption}
\usepackage{mathptmx}
\usepackage{type1cm}
\usepackage[T1]{fontenc}

\def\BibTeX{{\rm B\kern-.05em{\sc i\kern-.025em b}\kern-.08em
    T\kern-.1667em\lower.7ex\hbox{E}\kern-.125emX}}

\makeatletter
\@ifundefined{IEEEkeywords}{%
  \newenvironment{IEEEkeywords}{%
    \par\noindent\textbf{Keywords: }\itshape
  }{\par}%
}{}
\makeatother

\title{\LARGE \bf
GUIDER: Evaluating Goal-Free Human Intent Inference for Teleoperated Manipulation on Real-Robot Data
}

\author{Nicholas Kenny$^{1}$\orcidlink{0009-0008-2726-2576}, Cesar Alan Contreras$^{2}$\orcidlink{0009-0005-2866-8672}, Basile Ouedraogo$^{1}$\orcidlink{0009-0002-1398-4005},\\Rustam Stolkin$^{2}$, Manolis Chiou$^{3, *}$\orcidlink{0000-0002-9779-4067}, and Maria Kyrarini$^{1,4}$\orcidlink{0000-0003-0968-2477}
\thanks{$^{*}$Corresponding author: Manolis Chiou, {\tt\small m.chiou@qmul.ac.uk}}%
\thanks{$^{1}$ Human--Machine Interaction and Innovation Lab ($HMI^2$), Dept. of Electrical \& Computer Engineering, Santa Clara University, Santa Clara, CA, USA.}
\thanks{$^{2}$ Extreme Robotics Lab, School of Metallurgy and Materials, University of Birmingham, Birmingham, United Kingdom.}
\thanks{$^{3}$ School of Electronic Engineering and Computer Science, Queen Mary University of London, London, United Kingdom.}
\thanks{$^{4}$ Intelligent Systems Engineering  Dept., Luddy School of
Informatics, Computing, and Engineering, Indiana University, Bloomington, IN, USA}
}

\begin{document}

\maketitle
\thispagestyle{empty}
\pagestyle{empty}


\begin{abstract}
This paper presents an evaluation of a goal-free probabilistic framework for human intent inference during robotic manipulation. We deploy the Global User Intent Dual-phase Estimation for Robots (GUIDER) on data collected from a robotic arm to test the manipulation phase across various assistance scenarios, including making tea and fetching medicine. To support operation, we add online probability updates, workspace limits, support-plane filtering, and a grasping mode that prioritizes feasible grasp regions, all of which are tested on the recorded data while preserving its original temporal conditions. Across 20 manipulation steps in three scenarios, GUIDER estimated human intent within the correct grasp-candidate set in all cases, and achieved a time to confident prediction of 3.7 s, a remaining time before first grasp of 49.6 s, a prediction stability of 96.4\%, and a runtime of 4.857/4.474 s (mean/median) per perceptual phase of intent.
\end{abstract}


\section{Introduction}

Telepresence through remote teleoperation systems enables humans to control robotic manipulators in remote environments, such as hazardous industrial and nuclear facilities, disaster-response zones, underwater and space environments, healthcare settings, assistive robotic settings, and other remote manipulation tasks where full autonomy is difficult or undesirable \cite{10841491}. However, effective teleoperation remains challenging because operators must continuously translate high-level intentions into low-level control commands while maintaining situational awareness of a remote environment. This process often increases cognitive workload, reduces task efficiency, and can negatively affect human--robot teaming performance \cite{odoh_teleop_workload_2024}.


Recent telepresence research has therefore focused on integrating shared autonomy to reduce operator burden while preserving human control over task execution \cite{10841690}. Human intent inference is especially important in high-consequence domains where operator workload, safety, and task success are coupled. In medical and assistive robotics, understanding human intent and desires is essential for providing effective assistance \cite{jain_2019}. In nuclear decommissioning, anticipating operators' intent improves teleoperation performance and safety by maintaining low operator cognitive load in hazardous, complex situations \cite{pulgarin_2022}. In industrial scenarios, intention recognition augments the productivity of shared-autonomy frameworks by increasing task fluency and safety \cite{kekana_2025}.

\begin{figure}[h]
    \centering
    \includegraphics[width=0.49\textwidth]{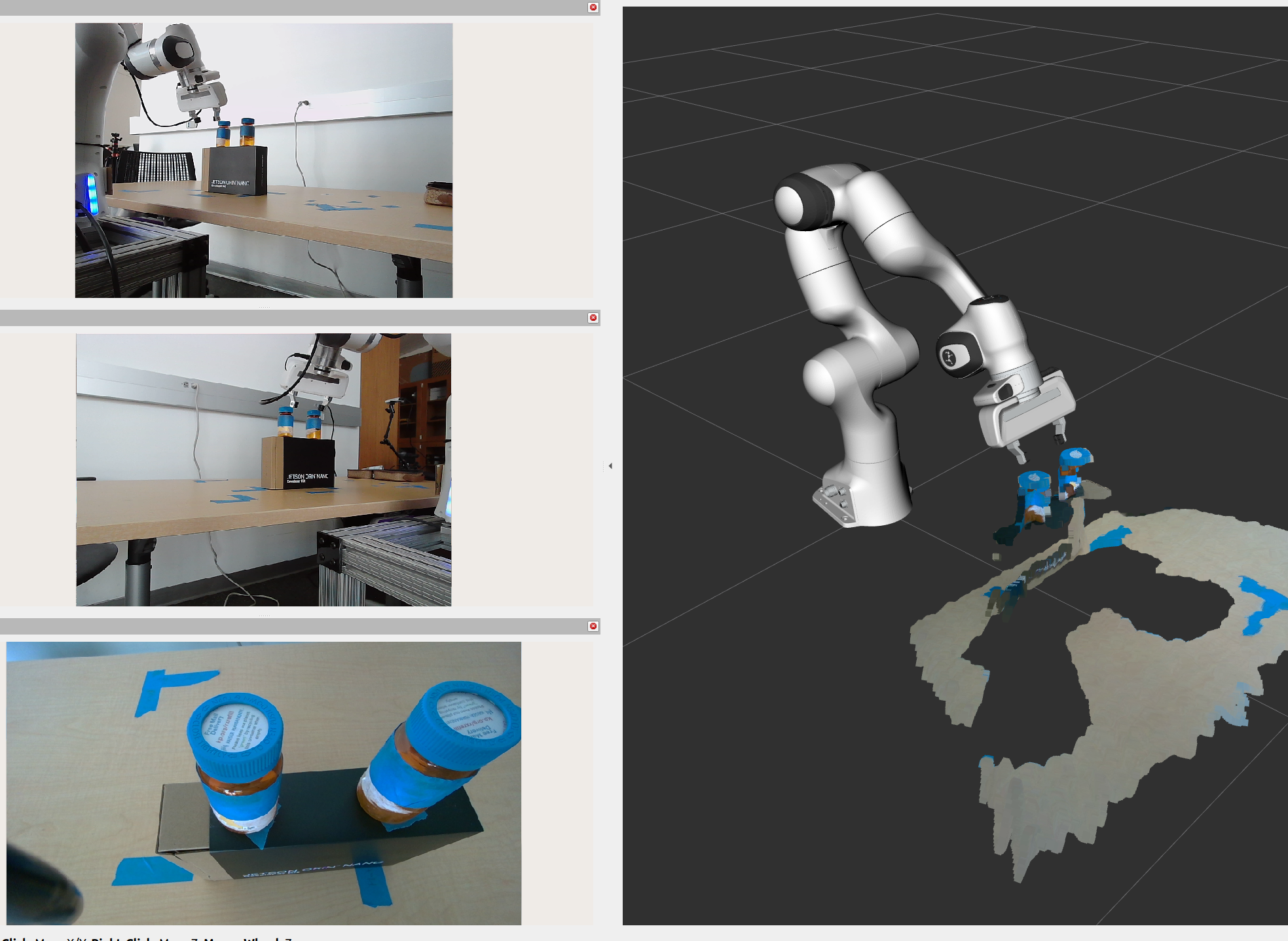}
    \caption{Operator Interface in RViz}
    \label{fig:op-interface}
\end{figure}

Our previous work introduced the Global User Intent Dual-phase Estimation for Robots (GUIDER), a dual-phase, goal-free probabilistic framework that predicts area-level navigation intent and object-level manipulation intent within a coupled belief structure. It was evaluated in Isaac Sim during human teleoperation trials \cite{contreras_2025}. In this study, we investigate whether those simulation results transfer to real-robot sensing and control conditions using temporally faithful replay of recorded teleoperation trials. Real-world data were collected on a physical system comprising a manipulator and a stereo depth camera. We developed an interface (see Fig. \ref{fig:op-interface}) that enables users to control the robot remotely without having to look directly at it. For safety during data collection, we used conservative speed and acceleration limits, enabled self-collision checking, and added controller gating to prevent unsafe commands. Finally, we manually calibrated the RGB-D camera so that the object and robot frames remained aligned, which was necessary for reliable perception and grasp-feasibility checks during replay. GUIDER was then evaluated on the recorded RGB-D and motion streams across three teleoperation task scenarios while preserving the original timing of each trial. 



The work presented in this paper contributes towards the deployment of intent inference in physical robots as follows: \textbf{First}, we extend GUIDER with several deployment-relevant contributions. These include online probability updates, limiting the sensed workspace to the robot's reach, removing points near the support plane before saliency, segmentation, and grasp checks, and adding a grasping mode that focuses on graspable regions instead of only object regions. \textbf{Second}, we evaluate GUIDER and the proposed improvements through temporally faithful replay of recordings collected on a physical robot across three scenarios, demonstrating how the method moves from simulation to the real world. \textbf{Third}, we report the main practical issues and engineering choices needed to build the real-robot collection and replay pipeline, including driver compatibility, real-time kernel placement, safety limits, and calibration.

\section{Related Work}
Previous related work can be grouped into three categories: (A) probabilistic inference of intentions, (B) shared autonomy in teleoperation, and (C) multimodal and interaction-sensitive modeling.
 
\subsection{Probabilistic Inference of Intention}
A recurring trend in manipulation-intent research is the shift from deterministic target prediction to uncertainty-aware representations that remain useful in the presence of ambiguous user behavior. Song et al.  \cite{song2024robottrajectron} exemplify this shift with Robot Trajectron, which predicts a distribution over future end-effector motion from recent kinematic history and uses that forecast to support shared-control manipulation. Zhao et al. \cite{zhao2024conformalized} extend this perspective by showing that uncertainty must also be calibrated when mapping low-dimensional teleoperation inputs to high-dimensional robot actions, so that the system can recognize when an inferred action is unreliable rather than overcommit to it. Shao et al. \cite{Shao2024} move one step further by combining particle-filter-based intent estimation with robot kinematic constraints and human manipulability measures, showing that likely intent and physical feasibility should be reasoned about jointly in co-manipulation.

Taken together, these studies suggest that robust manipulation-intent inference requires more than predicting what the user is likely to do; it must also explicitly represent confidence and relate that confidence to executable action hypotheses. GUIDER follows this direction by maintaining a goal-free belief over candidate grasp targets, updated online from scene perception and end-effector motion, and refined through grasp-feasibility constraints. In this sense, GUIDER is not separate from this line of work, but a continuation of it toward grasp-target inference that is both uncertainty-aware and physically actionable.


\subsection{Shared Autonomy and Implicit Intent Estimation}
If probabilistic intent estimates are to be useful in practice, they must ultimately inform how and when the robot assists. This is the central concern of shared-autonomy research, where intent estimation is often embedded directly inside the assistance policy rather than exposed as an explicit, reusable belief state. Zurek et al. \cite{zurek2021situational} address this issue through situational confidence assistance, in which the robot monitors whether its current repertoire of known intents can still explain the user's behavior and returns control to the user when confidence drops. Bowman et al.  \cite{bowman2024intent} present an intent-based task-oriented shared-control framework for telemanipulation, where inferred task intent is used to generate grasp poses and blend user motion with task-completion constraints. Belsare et al. \cite{belsare2025zeroshot} extend this direction with a zero-shot, vision-only shared-autonomy framework for tabletop manipulation, showing that intent recognition can be performed without relying on a fixed closed set of predefined targets. 

These systems show that early intent estimation becomes valuable when it can shape assistance, arbitration, and control authority during execution. However, in most of them, the inference module remains tightly coupled to the shared-control policy. GUIDER instead focuses on the inference layer itself, producing a structured probability distribution over feasible grasp candidates that can later support different shared-control strategies and applications. This distinction is important because it separates estimating what the user is likely trying to do from deciding how the robot should intervene.

\subsection{Multimodal and Interaction-sensitive Modeling}
Once intent estimates are expected to guide assistance during manipulation, motion cues alone are often insufficient. The robot must also reason about scene context, user interaction signals, and execution constraints. This motivates multimodal and interaction-sensitive approaches, in which intent is inferred from combinations of movement, gestures, objects, and spatial relationships. Oh et al. \cite{Oh2021}, for example, use hand-gesture information in teleoperation to predict user intent during manipulation tasks. Related work on spatial intention maps further shows that explicitly representing relationships among agents, objects, and goals can improve context-aware reasoning in mobile manipulation settings \cite{Wu2021}. Shao et al. \cite{Shao2024} extend this line of work by showing that task structure, environment context, and physical constraints can be incorporated jointly in collaborative object manipulation. These studies collectively show that robust intent inference depends on combining perceptual cues with contextual reasoning rather than relying on motion alone. At the same time, many approaches like these still emphasize either prediction or contextual perception in isolation, and physical feasibility is often only partially integrated into the inference process.

GUIDER follows this multimodal direction by integrating RGB-D perception, saliency, segmentation, geometric grasp checks, and end-effector motion within one probabilistic update loop. Rather than treating perception, intent estimation, and feasibility as separate stages that are only loosely connected, GUIDER maintains a structured belief over candidate grasp targets and refines that belief with visual and kinematic evidence as interaction unfolds. This allows the system to support shared-autonomy policies while keeping the inferred intent physically possible under real-world manipulation constraints.


\section{System Overview}
GUIDER is a dual-phase, goal-free probabilistic intent inference framework that maintains a coupled belief over navigation intent and manipulation intent \cite{contreras_2025}.  In the original system, the navigation phase uses controller motion and map context to estimate the likely interaction area, and once that area is reached, the manipulation phase shifts the belief to object-level intent. The manipulation phase combines RGB-D perception with U$^2$-Net saliency, FastSAM instance segmentation, and 3D geometric analysis (Random Sample Consensus (RANSAC) and Hierarchical Density-Based Spatial Clustering of Applications with Noise (HDBSCAN)) to produce object hypotheses from the local scene. These hypotheses are then filtered by grasp-feasibility tests that encode whether the current end-effector can meaningfully approach and grasp each candidate. Finally, candidate probabilities are updated online based on end-effector motion, so objects that remain consistent with the operator's approach gain belief, while inconsistent candidates decay.

\subsection{Implementation of GUIDER}
In this paper, we evaluate the manipulation phase of GUIDER on data recorded from a real robot. The navigation phase from \cite{contreras_2025} is part of the original framework, but it is not evaluated here, and no mobile-base intent calculations are used in the experiments. We keep the main GUIDER pipeline, but tune it for live robot operation for manipulation and prepare the system to provide user assistance (e.g., shared control) in future iterations. Below, we discuss the explicitly modified parts of the GUIDER manipulation phase.

Let
\[
\mathcal{D}_{t_0}=\{\mathbf{I}_{t_0},D_{t_0},\mathbf{q}_{t_0},T_{t_0}\}
\]
denote the RGB observation \(\mathbf{I}_{t_0}\), depth observation \(D_{t_0}\), end-effector state \(\mathbf{q}_{t_0}\), and required frame transforms \(T_{t_0}\) available at the GUIDER intent trigger time \(t_0\). The static manipulation-perception pipeline computes the retained candidate set and its initial probability state as
\[
(\hat{\mathcal{O}}_{t_0},\hat{\mathbf{p}}_{t_0})
=
\mathcal{F}_{\mathrm{static}}(\mathcal{D}_{t_0}),
\]
where \(\mathcal{F}_{\mathrm{static}}\) denotes the GUIDER static pipeline used to construct the manipulation candidates.

The active candidate set is
\[
\hat{\mathcal{O}}_{t_0}=\{o_k\}_{k=1}^{N_{t_0}},
\qquad
N_{t_0}=|\hat{\mathcal{O}}_{t_0}|,
\]
where \(N_{t_0}\) is the number of retained active candidates at trigger time. In Object Mode, \(o_k\) denotes an object candidate. In Grasping Mode, \(o_k\) denotes a feasible grasp-region candidate. The corresponding initial candidate-level probability state is
\[
\hat{\mathbf{p}}_{t_0}
=
[p_1(t_0),\ldots,p_{N_{t_0}}(t_0)]^\top .
\]

The manipulation intent perception pipeline of GUIDER maintains its formulation in the image domain, with the RGB observation denoted by
\[
\mathbf{I} : \Omega \subset \mathbb{Z}^2 \to \mathbb{R}^3,
\]
and the aligned depth image by
\[
D : \Omega \subset \mathbb{Z}^2 \to \mathbb{R}_{\ge 0},
\]
where \(\Omega\) is the image lattice and each pixel \((u,v) \in \Omega\) is associated with a color vector \(\mathbf{I}(u,v)\) and a depth value \(D(u,v)\). Given camera intrinsics \((f_x,f_y,c_x,c_y)\), each valid pixel can be lifted to 3D through the standard back-projection map

\[
\mathbf{X}(u,v) =
\begin{bmatrix}
\frac{(u-c_x)D(u,v)}{f_x} \\
\frac{(v-c_y)D(u,v)}{f_y} \\
D(u,v)
\end{bmatrix}.
\]

The corresponding back-projected point set is
\[
\mathcal{P}=\{\mathbf{X}(u,v)\in\mathbb{R}^3 \mid (u,v)\in\Omega,\ D(u,v)>0\}.
\]

This representation defines the geometric support used by RANSAC, HDBSCAN, graspability analysis, and subsequent 3D evolving probability. In the evaluation, we also apply a distance truncation to discard points beyond 1 meter. This threshold is driven by the effective reach of the manipulator and the operating camera field of view during near-field telemanipulation. Thus, if \(\mathbf{X}_i\in\mathcal{P}\) is a back-projected point in the camera frame, it is retained only if
\[
\mathcal{P}_{\text{kept}} = \{\mathbf{X}_i \in \mathcal{P} \mid \|\mathbf{X}_i\|_2 \le 1.0\}.
\]

A second deployment change removes points that are too close to the dominant RANSAC support plane. This suppresses artifacts from surface texture and small geometric irregularities (e.g. paper labels, barcodes, or furniture bumps) that otherwise create unstable false positives. After estimating the dominant plane, the system computes the pixel distance to that plane and removes evidence below a clearance threshold.

Let the dominant plane be
\[
\pi: \mathbf{n}^\top \mathbf{X} + d = 0,
\]
with normal \(\mathbf{n}\) and offset \(d\). For each valid pixel \((u,v)\), back-projection yields a 3D point \(\mathbf{X}(u,v)\). The absolute plane distance is then
\[
\delta_\pi(u,v) = \frac{|\mathbf{n}^\top \mathbf{X}(u,v) + d|}{\|\mathbf{n}\|_2}.
\]
The current system defines a binary plane-clearance mask
\[
M_\pi(u,v) =
\begin{cases}
1, & \delta_\pi(u,v) \ge \tau_\pi,\\
0, & \text{otherwise},
\end{cases}
\qquad \tau_\pi \in [0.01, 0.03]\,\text{m}.
\]
The clearance threshold \(\tau_\pi\) was selected per scenario within this range to suppress camera-depth noise near the support surface while preserving valid object evidence.

This mask is applied throughout the perception pipeline. If \(S(u,v)\) is the U2Net saliency map and \(F(u,v)\) is the FastSAM mask, the implementation enforces
\[
S'(u,v) = S(u,v)\, M_\pi(u,v),
\]
\[
F'(u,v) = F(u,v)\, M_\pi(u,v).
\]

Thus, the fused probability map becomes
\[
P(u,v)=\Psi(S',F',\ldots)(u,v),
\]
followed by a final hard gating with the same \(M_\pi\). In practice, the support plane acts as an explicit exclusion set. The three geometric grasp-feasibility tests from prior GUIDER are unchanged; now they operate on saliency/segmentation evidence that has already been plane-filtered through \(M_\pi\).

The deployment implementation also maintains an active intent record and updates candidate probabilities online as the end-effector moves. During one triggered inference episode, the active candidate set, \(\hat{\mathcal{O}}_{t_0}\), is held fixed and the probability of each retained candidate evolves as
\[
p_k(t+\Delta t)=\mathcal{U}\big(p_k(t),\mathbf{x}^{\mathrm{eef}}_{t+\Delta t},o_k\big),
\qquad k=1,\ldots,N_{t_0},
\]
where \(\mathbf{x}^{\mathrm{eef}}_t\in\mathbb{R}^3\) is the end-effector position and \(\mathcal{U}\) is the runtime probability-evolution operator (for a more in-depth explanation, see \cite{contreras_2025}).

Finally, we introduce a Grasping Mode. In deployment, this mode replaces Object Mode during manipulation operations, so assistance focuses on actionable grasp regions instead of only object localization. Unlike Object Mode, where the final fused map represents object-like regions, Grasping Mode transforms that support into grasp-centric affordance regions derived from the advanced rectangle mask from GUIDER \cite{contreras_2025}.

Let \(A_{\text{rect}}(u,v)\) be the advanced graspability rectangle mask. A morphological closing operator \(\mathcal{C}\) is applied to obtain a more stable grasp-region mask
\[
G(u,v) = \mathcal{C}\big(A_{\text{rect}}(u,v)\big).
\]
The fused probability map \(P(u,v)\) is then hard-gated:
\[
P_{\text{grasp}}(u,v) = P(u,v)\, G(u,v).
\]

As a result, extracted connected regions are no longer generic object extents, but action-conditioned feasible grasp regions. In Object Mode, the system estimates ``what object is present and likely relevant." In Grasping Mode, it estimates ``where the tool can meaningfully act." This is a deliberate shift from semantic objectness to operational affordance, designed to support meaningful assistance with the current two-finger end-effector. This GUIDER configuration is evaluated in Sec. \ref{sec:experimental_protocol} across three environments.

\subsection{Control architecture}
The manipulator is a Franka Emika Panda arm (Panda for short) commanded through "MoveIt 2 Servo" package, using a unified robot description that includes both the manipulator and camera frames to support consistent transforms and accurate depth information \cite{moveit2_servo}. Arm control is performed using a custom integration of the ROS2 Humble \verb|teleop_twist_joy| and \verb|moveit_servo| packages, which allows for complete teleoperation control of the manipulator using a Thrustmaster T-Flight Stick X USB joystick \cite{teleop-twist-joy} \cite{moveit2_servo}. 
"MoveIt 2 Servo" maps joystick input to Cartesian tool-center-point motion; dedicated buttons control the gripper, switch between translational and rotational axes, and return the arm to the ready pose. For ease of repositioning the arm back to the ``ready" position, control can be toggled from the MoveIt servo controller to the default MoveIt pathplanner with just one button press on the controller; an additional button repositions the arm automatically, regardless of starting state \cite{moveit}. 

Perception streams provide RGB-D data for the manipulation phase. The RGB-D data is captured by an Intel RealSense D405 stereo depth camera mounted at the last joint of Panda above the 2-finger gripper. This particular camera was chosen because it is designed for short-range applications with high accuracy, such as robotic manipulation. 

\section{Experiment}
\subsection{Experimental Setup}
\begin{figure}[h]
    \centering
    \begin{minipage}{0.20\textwidth}
        \centering
        \includegraphics[width=\textwidth]{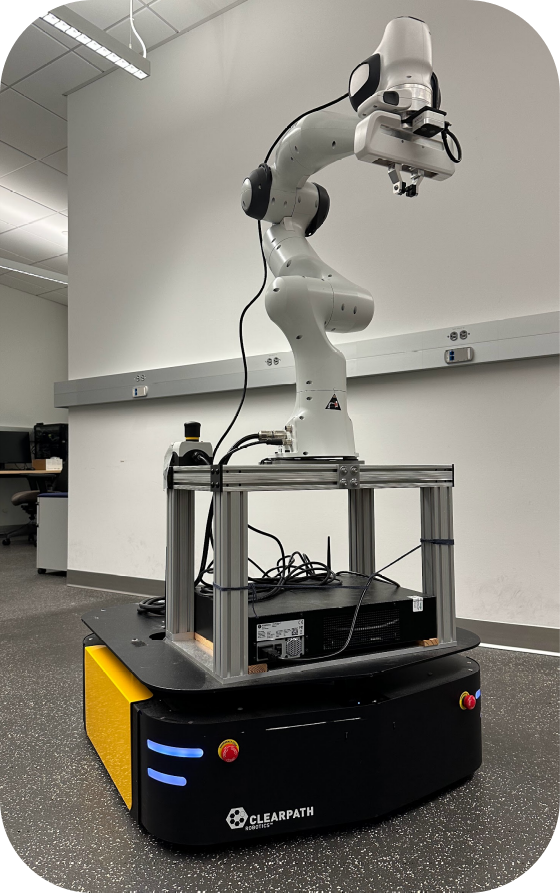}
        \caption{Clearpath Ridgeback with Franka Emika Panda Arm}
        \label{fig:deployment_and_explanation_1}
    \end{minipage}
    \hfill
    \begin{minipage}{0.24\textwidth}
        \centering
        \includegraphics[width=\textwidth]{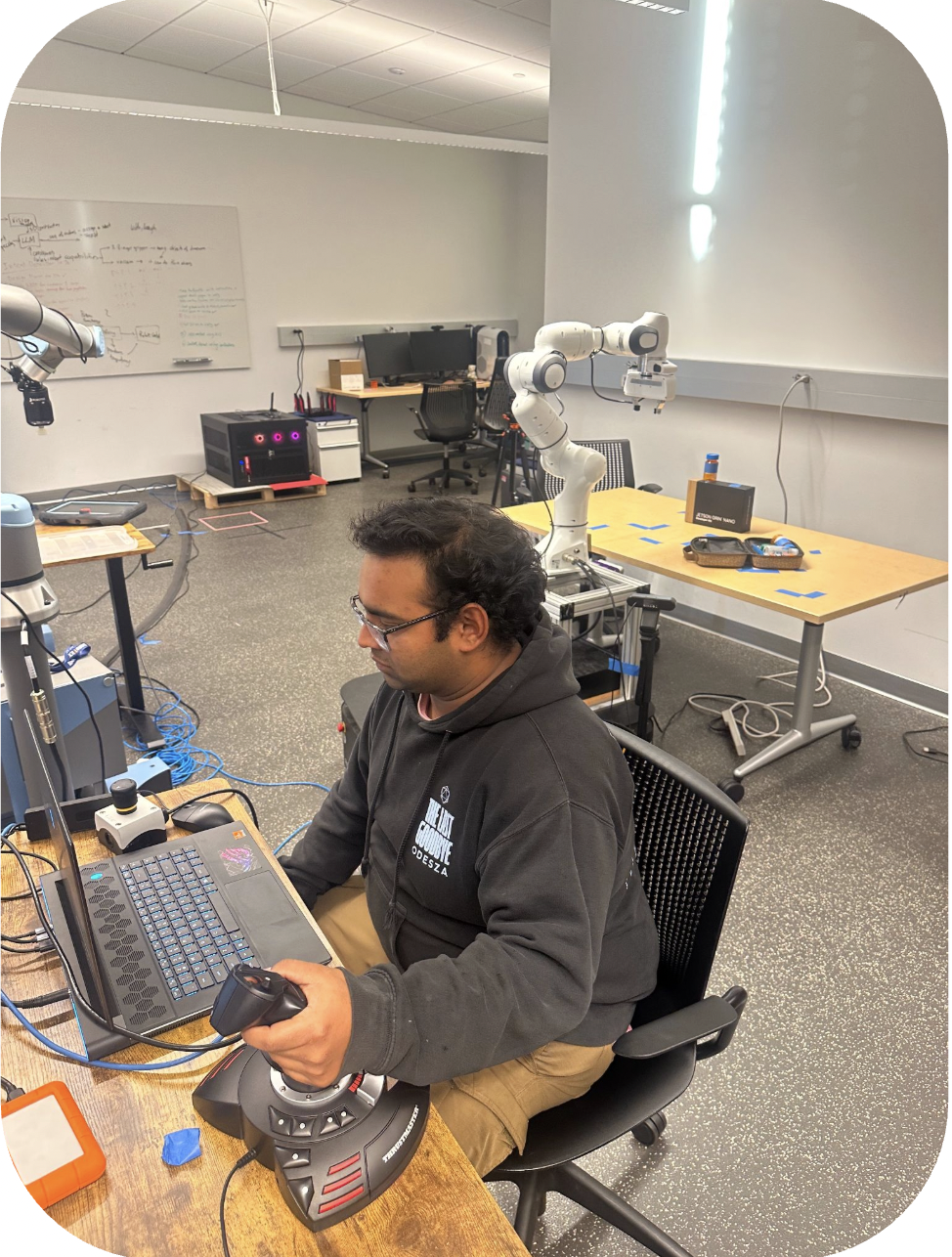}
        \caption{Operator shown controlling the Franka Emika Panda Arm}
        \label{fig:operator_and_arm_1}
    \end{minipage}
\end{figure}

Fig. \ref{fig:deployment_and_explanation_1} shows the configuration of the Panda used in this study. The manipulator and its controller are mounted to a Clearpath Ridgeback mobile base (which is not used in these experiments). Control of Panda manipulators requires a strict real-time connection between the workstation PC and the robot at a minimum of 1-kHz cycle time. To facilitate this control, we used a laptop running Ubuntu 22.04 LTS on the Linux 5.15 kernel with the \verb|PREEMPT-RT| patchset. The laptop was connected directly via Ethernet to the Panda controller, and communication was established using the FCI via ROS2.

As shown in Fig. \ref{fig:operator_and_arm_1}, the operator is positioned in a way such that they cannot directly see the manipulator and the task area. The only viewpoints the operator can use to gain situational awareness of the robot and task area are video streams from \textbf{a)} the two USB cameras positioned at table height to the left and right of the manipulator, and \textbf{b)} the end-effector camera. The video streams are displayed on a monitor in a simple visual interface (Fig. \ref{fig:op-interface}). The robot is controlled via a joystick while the operator is looking at the monitor. This setup mimics a scenario in which the operator is remotely controlling the manipulator from a distant location. 

\subsubsection{Integration constraints and engineering decisions}
The deployment of our data collection platform introduced a set of engineering constraints that required a set of engineering decisions for reliable operation. Our deployment provides practical guidance for future implementations of similar systems. \textbf{First}, stable arm control required strict compatibility between the Franka driver and the libfranka version used by the ROS2 Humble stack \cite{libfranka}. The arm we used is the Franka Emika Robot (FER), which is not natively compatible with \verb|franka_ros_2| and instead required us to use \verb|multipanda_ros2| with \verb|libfranka| version 0.9.2 \cite{skerlj2026multipanda_ros2} \cite{libfranka}. \textbf{Second}, "MoveIt 2 Servo" operation was sensitive to timing jitter, so arm control was placed on a real-time kernel to reduce scheduling variability. Additionally, we used a basic low-pass filter to reduce arm jitter during joystick teleoperation. \textbf{Third}, the operator perspective using a single camera mounted at the end-effector was extremely limited. To resolve this issue, we added two separate USB webcams to provide additional views of the scene and improve teleoperation. \textbf{Finally}, perception quality depended on accurate sensor-frame alignment, so RGB-D calibration was performed manually to maintain stable transforms required by perceptual inference in GUIDER.

\subsection{Experimental Protocol} \label{sec:experimental_protocol}
For each experiment, data was gathered from a single human operator\footnote{This evaluation was designed as a technical transfer study. The single-operator results characterize system feasibility under the tested conditions.}. The operator received standardized training on the robot controls and task procedure, and was given a series of steps for the items he should move in each scenario. Before beginning each step in a scenario, the operator moved the robot to a comfortable pose for the current task and held it stationary for 10 seconds to allow the initial 3D reconstruction and manipulation-perception pipeline to stabilize. The operator's time to first grasp attempt was measured from the moment the robot began moving after the idle time. During all trials, the full ROS2 data was recorded, including robot motion, RGB-D observations, and relevant system states. After data collection, the recordings were replayed while preserving their original temporal evolution, allowing GUIDER to be evaluated on the same sensing and motion sequence observed during the live experiment.

\subsubsection{Scenarios}
GUIDER was evaluated in three teleoperation scenarios to test whether the method would still work as the scene, object arrangement, and task demands changed. These scenarios allow us to study how well the system transfers to real-world settings with varying levels of precision, object similarity, and other variations. During data collection, no shared-autonomy assistance was active, and the operator controlled the manipulator directly across all scenarios.

\begin{figure}[h!]
  \centering
  
  \begin{subfigure}[b]{0.22\textwidth}
    \centering
    \includegraphics[width=\textwidth]{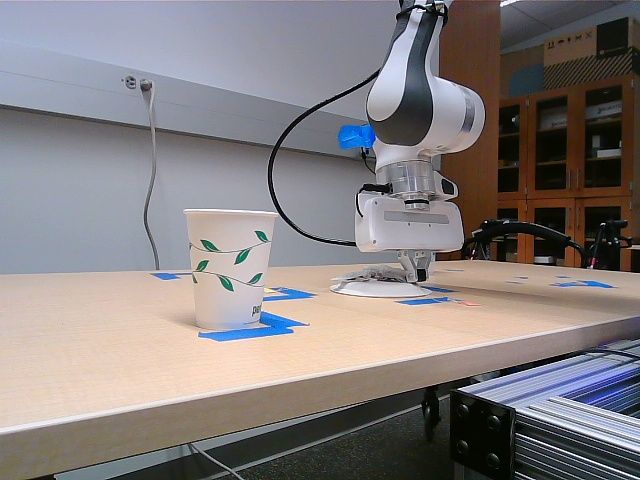}
    \caption{Grasping tea bag}
    \label{fig:tea}
  \end{subfigure}
  \hfill
  \begin{subfigure}[b]{0.22\textwidth}
    \centering
    \includegraphics[width=\textwidth]{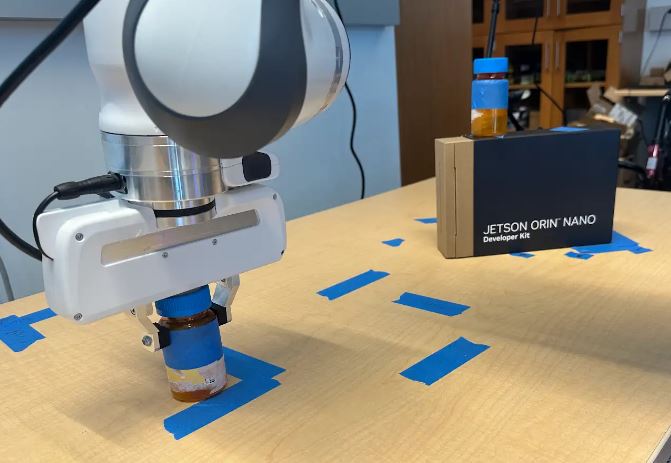}
    \caption{Fetching Medicine}
    \label{fig:medicine}
  \end{subfigure}


  \begin{subfigure}[b]{0.22\textwidth}
    \centering
        \includegraphics[width=\textwidth]{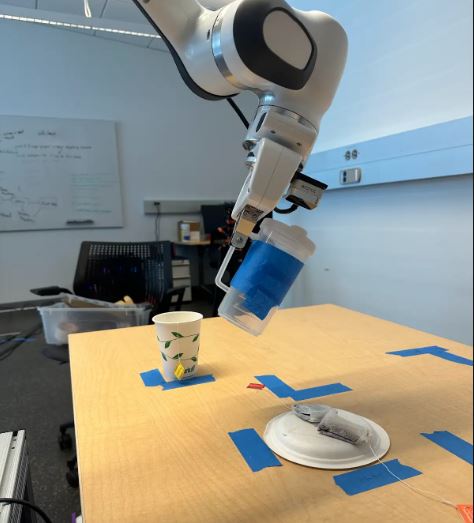}
    \caption{Grabbing water pitcher}
    \label{fig:pitcher}

  \end{subfigure}
  \hfill
  \begin{subfigure}[b]{0.22\textwidth}
    \centering
    \includegraphics[width=\textwidth]{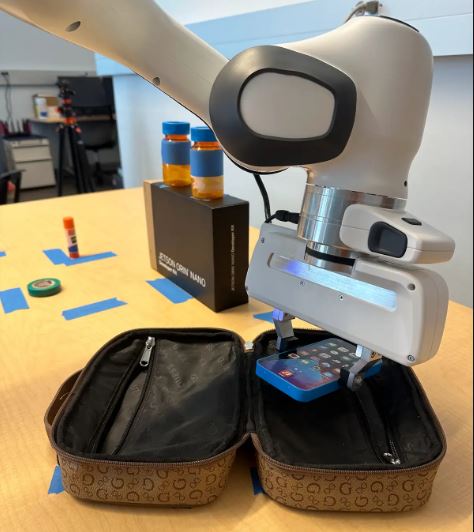}
    \caption{Handling a belongings bag}
    \label{fig:belongings}
  \end{subfigure}

  \caption{Robot performing tasks under the three scenarios}
  \label{fig:tasks}
\end{figure}

\emph{\textbf{Scenario 1:}} Making Tea (Figs \ref{fig:tea}, \ref{fig:pitcher}). The task starts with selecting and grasping one tea bag from a set of three similar tea bags, then dropping it into a container. The operator then grasps a water container and pours water into the teacup. For this scenario, the main points of interest are the grasping steps for the tea bag and the water container. This scenario is especially interesting for GUIDER because the tea bags are small and visually similar, whereas the water container has a more complex geometry, making grasp selection more important.

\emph{\textbf{Scenario 2:}} Fetching Medicine (Fig. \ref{fig:medicine}). The task consists of two medicine containers placed on a table. The packs look similar and are close together, but the operator must grasp the correct one and place it in a container. This scenario is especially interesting for GUIDER because it tests whether the system can separate similar nearby objects and guide attention toward the correct grasp target.

\emph{\textbf{Scenario 3:}} Handling belongings (Fig. \ref{fig:belongings}). A set of different belongings is scattered across a table. The objects have different shapes, sizes, and colors. The operator must grasp them, place them in a bag, and then close the bag. For this scenario, the main point of interest is whether GUIDER can still work well when the target objects are varied and visually different, rather than similar to each other.

\subsubsection{Tasks and metrics}
As in our previous work, we report prediction timing and stability, but we also include two practical real-world measures related to grasping performance:
\\
\textbf{Time to confident prediction (TCP):} Time from the GUIDER evolving probability phase start to the first sustained correct prediction of the true grasp target.
\\
\textbf{Remaining time at confident prediction (RTCP):} Time between the first correct prediction from GUIDER and the operator's first grasp attempt on the object of intent.
\\
\textbf{Prediction stability:}  Remaining time interval after TCP during which the predicted target remains correct, divided by the total time of the interval.
\\
\textbf{Target within predicted items:} Whether the desired target is included in the full set of items predicted by GUIDER as grasp candidates. This tells us whether the correct target remains among the predicted candidate items, even when the final ranking within that set is not yet fully settled.
\\ 
\textbf{Human time to first grasp:} Time from when GUIDER starts computing predictions to the operator's first grasp attempt on the target object. This provides a real-world reference for how long it takes a human to act on a target.
\\
\textbf{Runtime:} Time of the perceptual manipulation phase of GUIDER.

\section{Results}
Table \ref{tab:trial_metrics} shows that GUIDER kept the true target within the predicted grasp-candidate set in all 20 evaluated manipulation steps across the three scenarios. Across all steps, the mean TCP was 3.7s (median 3.0s), and the mean RTCP was 49.6s (median 41.8s), indicating that the correct target was typically identified well before the operator attempted the first grasp. In 19 of the 20 steps, TCP occurred within the first 8s of the interaction. The only later case was the bag-handle step in the belongings scenario (17.7s). Prediction stability was high overall, with a mean of 96.4\%, and 14 of the 20 steps maintained a correct prediction for the entire remaining interval after TCP.

Performance remained consistent across scenarios, although task structure and scene complexity affected the time margins on the perceptual side, due to the number of grasping candidates. Making Tea produced the largest average RTCP (58.4s), mainly because the water-container handle was hard for the human controlling the robot to grasp, while it was a salient zone easy for GUIDER to identify.

Fetching Medicine showed the lowest average perception runtime (3.9s) while maintaining a mean prediction stability of 96.7\%, because the medicine containers were highly salient. Finding Belongings was the most perceptually demanding case, with the highest average runtime (6.1 s), but still retained the correct target in the predicted set for every object and achieved a mean stability of 98.6\%. The lowest stability was observed for the third tea-bag step (67.4\%), indicating that small, visually similar objects are the hardest case for the current GUIDER perception pipeline, especially when clutter introduces many competing salient regions.

The runtime breakdown in Table \ref{tab:GUIDER_stages} shows that the deployed manipulation phase runs with a mean/median computation time of 4.857/4.474 s per intent update. FastSAM filtering is the dominant component (2.007/1.894 s), followed by HDBSCAN clustering (0.863/0.813 s), 2D-to-3D reconstruction (0.671/0.745 s), and RANSAC plane fitting (0.421/0.354 s). Together with U$^2$-Net saliency, these stages account for most of the runtime budget, whereas the online motion-probability evolution step is negligible at 0.000145/0.000125 s per update. This shows that, in the real deployment, runtime is dominated by perception rather than by the belief-update mechanism itself.

\begin{table}[h]
\centering
\begingroup
\scriptsize
\caption{Main stages timing of the manipulation phase of the GUIDER framework. Reported values are mean/median runtime (Across 20 intent detections)}
\label{tab:GUIDER_stages}
\setlength{\tabcolsep}{4pt}
\renewcommand{\arraystretch}{1.05}
\begin{tabular}{p{0.52\columnwidth} p{0.30\columnwidth}}
\hline
\textbf{Stage} & \textbf{Mean / Median [s]} \\
\hline
2D$\rightarrow$3D point cloud reconstruction & 0.671 / 0.745 \\
RANSAC plane fitting & 0.421 / 0.354 \\
HDBSCAN clustering & 0.863 / 0.813 \\
U$^2$-Net saliency detection & 0.370 / 0.404 \\
FastSAM filtering & 2.007 / 1.894 \\
Saliency probability fusion & 0.002839 / 0.002732 \\
Gripper bounding box check & 0.001788 / 0.001727 \\
Morphological graspability & 0.193 / 0.199 \\
Advanced graspability & 0.292 / 0.218 \\
Refined saliency combination & 0.010 / 0.009766 \\
Probability rescaling & 0.025 / 0.022 \\
\hline
Total GUIDER manipulation & 4.857 / 4.474 \\
\hline
Motion probability evolution (step) & 0.000145 / 0.000125 \\
\hline
\end{tabular}
\endgroup
\vspace{-1.0em}
\end{table}

\begin{figure*}[t]
  \centering
  \includegraphics[width=0.95\linewidth]{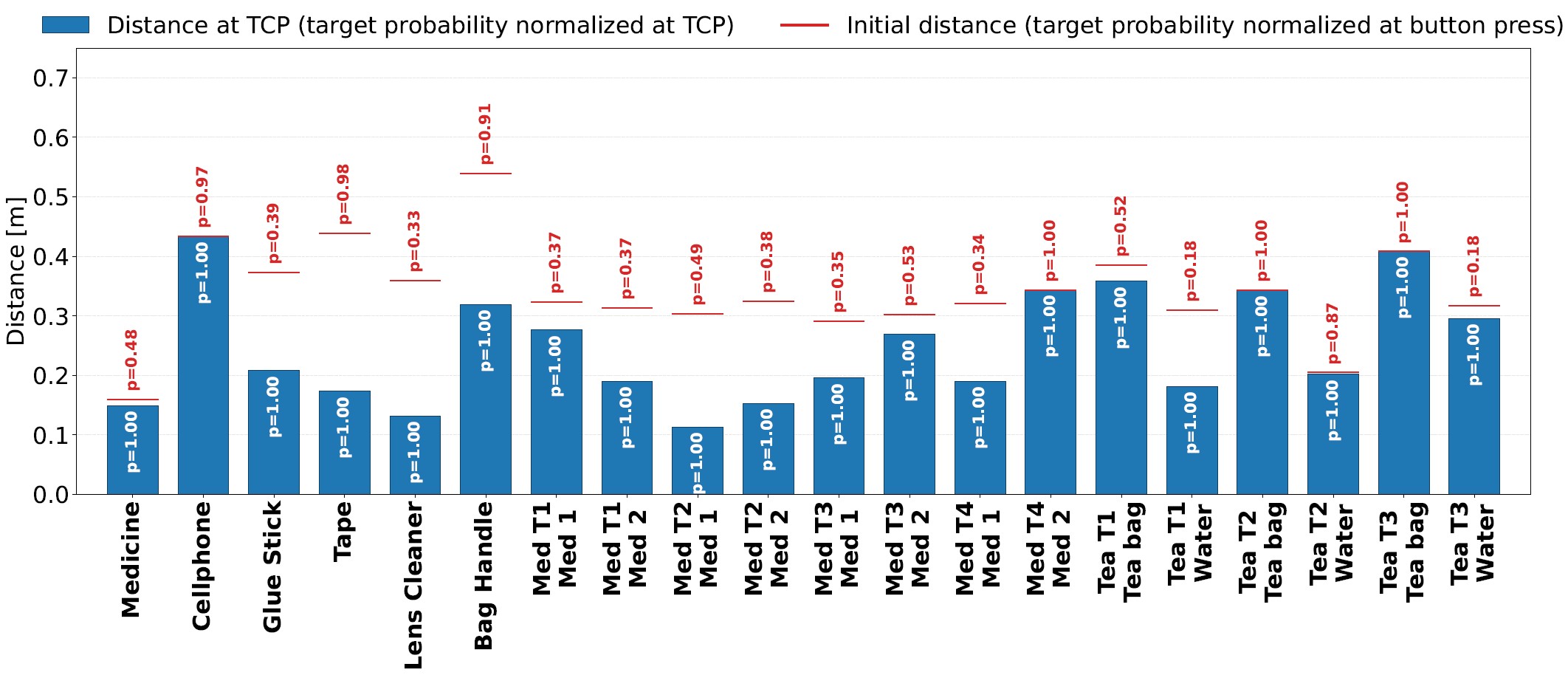}
  \caption{Initial and confident-prediction target distances across all real-world trials. Blue bars indicate the distance to the target at TCP, red markers indicate the initial distance at the start of GUIDER calculations, and the annotated probabilities are normalized by the maximum object probability at each respective time.
}
  \label{fig:pred_dist}
\end{figure*}

Fig. \ref{fig:pred_dist} complements the timing analysis with a geometric view of the interaction. Across all 20 steps, the mean end-effector-to-target distance decreased from 0.340 m at button press to 0.247 m at TCP, and 17 of the 20 steps reached TCP at a shorter target distance than the initial one. This suggests that confident prediction usually emerged after some approach motion, when motion and geometry cues had become more informative, while still leaving a substantial time margin for potential assistance. Because the core GUIDER formulation was evaluated and compared with various baselines previously in \cite{contreras_2025}, the present experiments are intended as an exploratory transfer study of the deployed manipulation pipeline.

\begin{table*}[t]
\centering
\caption{Per-trial results across the three scenarios. Making Tea and Fetching Medicine are reported step by step in each trial. For Finding Belongings, the single trial is reported object-by-object across the saved target objects. TCP is the time from the GUIDER evolving probability phase start to the first sustained correct prediction. RTCP is the remaining time from that prediction to the operator's first grasp action. Runtime reports the calculation time of the manipulation perception stage of GUIDER.}
\label{tab:trial_metrics}
\scriptsize
\setlength{\tabcolsep}{3pt}
\begin{tabular}{@{}lllcccccc@{}}
\toprule
Scenario & Trial & Step & TCP & RTCP & Prediction stability & Target within predicted items & Human time to first grasp & Runtime \\
\midrule
Making Tea & Trial 1 & Tea bag & 4.1s & 40.9s & 100.0\% & Yes & 45s & 5.52s \\
Making Tea & Trial 1 & Water container & 4.7s & 51.3s & 100.0\% & Yes & 56s & 5.21s \\
Making Tea & Trial 2 & Tea bag & 0.0s & 37.0s & 99.8\% & Yes & 37s & 4.60s \\
Making Tea & Trial 2 & Water container & 0.3s & 94.7s & 95.2\% & Yes & 95s & 3.44s \\
Making Tea & Trial 3 & Tea bag & 0.0s & 41.0s & 67.4\% & Yes & 41s & 5.49s \\
Making Tea & Trial 3 & Water container & 5.7s & 85.3s & 100.0\% & Yes & 91s & 4.91s\\
Fetching Medicine & Trial 1 & Med 1 & 6.5s & 20.5s & 100.0\% & Yes & 27s & 4.36s\\
Fetching Medicine & Trial 1 & Med 2 & 2.7s & 45.3s & 100.0\% & Yes & 48s & 4.14s\\
Fetching Medicine & Trial 2 & Med 1 & 7.3s & 47.7s & 86.2\% & Yes & 55s & 4.21s\\
Fetching Medicine & Trial 2 & Med 2 & 1.9s & 21.1s & 100.0\% & Yes & 23s & 3.95s\\
Fetching Medicine & Trial 3 & Med 1 & 3.3s & 39.7s & 100.0\% & Yes & 43s & 3.63s\\
Fetching Medicine & Trial 3 & Med 2 & 2.0s & 71.0s & 100.0\% & Yes & 73s & 3.75s\\
Fetching Medicine & Trial 4 & Med 1 & 2.3s & 75.7s & 87.4\% & Yes & 78s & 4.01s\\
Fetching Medicine & Trial 4 & Med 2 & 0.0s & 65.0s & 100.0\% & Yes & 65s & 3.13s\\
Finding Belongings & --- & Medicine & 0.5s & 42.5s & 91.8\% & Yes & 43s & 4.28s\\
Finding Belongings & --- & Cellphone & 1.1s & 33.9s & 100.0\% & Yes & 35s & 6.87s\\
Finding Belongings & --- & Glue Stick & 3.8s & 30.2s & 100.0\% & Yes & 34s & 5.45s\\
Finding Belongings & --- & Tape & 6.5s & 32.5s & 100.0\% & Yes & 39s & 7.69s\\
Finding Belongings & --- & Lens Cleaner & 3.3s & 38.7s & 100.0\% & Yes & 42s & 4.59s\\
Finding Belongings & --- & Bag Handle & 17.7s & 78.3s & 100.0\% & Yes & 96s & 7.91s\\
\bottomrule
\end{tabular}
\vspace{-2em}
\end{table*}

\section{Discussion}
The results show that GUIDER can be transferred from simulation data to real-world sensing data collected on a physical manipulation platform while preserving the main behavior of interest, an early and stable target inference. The combination of low TCP and high RTCP suggests that, once GUIDER identifies the correct grasp candidate, the belief is often available before the operator's first grasp action. This timing is relevant for future shared-autonomy studies in assistive robotics or hazardous materials handling, where an assistance layer (e.g., a shared autonomy layer) could use such a belief, if the perception latency is acceptable, to aid a human. The distance trends in Fig. \ref{fig:pred_dist} support this deduction, since TCP was usually reached after the operator had begun moving toward the object, indicating that GUIDER was not guessing from the initial scene alone while still leaving enough distance to assist the user.

The addition of the grasping mode also makes the output more directly aligned with the system's intended use. The system can now infer a meaningful manipulation action target (a specific grasp) rather than focusing on moving to the center of the object. GUIDER now focuses on the highest-probability saliency at grasp locations, rather than going directly to the center of an object.

The scenarios also show where the deployment is most sensitive. Finding Belongings produced the highest perception runtimes, which is consistent with the increased clutter and object diversity in that environment. Yet, once GUIDER reached a high prediction value, confidence remained highly stable. In contrast, the lowest stability occurred in the tea-bag condition, where the targets were small (2-3 cm height), close together, and visually similar. Small objects near the support plane are a broadly difficult RGB-D manipulation setting, but in our case the sensitivity arises specifically from the current combination of saliency estimation, instance segmentation, plane filtering, and clustering, which becomes less discriminative under plane noise and heavy visual similarity. This suggests that the size and placement of candidate objects remain a hard case for the present deployment and a clear target for future improvement.

From a systems perspective, the main bottleneck is clearly the perception stack. FastSAM alone contributes the largest share of the total runtime. Clustering and point-cloud processing add most of the remaining cost. By comparison, the online motion-probability evolution is effectively negligible and can be fully updated in real-time. This indicates that future optimization efforts should focus on segmentation and clustering latency rather than on the probabilistic update rule. In other words, the causal online formulation is computationally cheap, but the cost of GUIDER comes from constructing the perceptual evidence that feeds it. Before starting a new task, a human operator studies the environment for a few seconds, enough time for GUIDER to create an initial confident prediction. By informing the operator of this behavior prior to operation, we think GUIDER can be minimally disruptive to an operator's nominal workflows.

The paper demonstrates the feasibility of deployment with real-world sensing data and cross-scenario transfer, but it does not yet evaluate a closed-loop assistance policy that acts on GUIDER's predictions, nor a full online deployment. The current evidence supports portability and runtime feasibility. A natural next step is to connect the inferred belief state to assistance behaviors and measure whether the early predictions reported here translate into improved task fluency, reduced workload, or higher grasping efficiency.

\section{Conclusion}
This paper presented an evaluation of the GUIDER manipulation phase on data collected from a Franka Emika Panda robot across three representative scenarios. In addition to validating the transfer from simulation to real-world sensing data collected on a physical platform, the work introduced technical improvements that enabled this evaluation. The most important one is the grasping mode that converts object-centered evidence into feasible grasp regions. This shifts GUIDER from a purely object-centered manipulation predictor toward a grasp-oriented inference process that is better suited to real robot perception streams and future shared-autonomy use.

The results show that the pipeline can maintain the correct target within the predicted candidate set across all evaluated steps, generate confident predictions early relative to human grasping time, and operate with a mean calculation time of 4.857 s while keeping the online motion-based probability evolution effectively negligible in cost. Together, they also support the practical integration of the proposed modifications under the tested conditions.

Furthermore, the study identified the conditions required for the practical constraints relevant to eventual deployment, albeit in the form of a data collection pipeline. Future work will integrate GUIDER's online belief state into closed-loop shared-autonomy behaviors and evaluate whether the prediction margins reported here translate into measurable gains in assistance quality, efficiency, and operator workload.

\section*{Acknowledgments}
The authors would like to thank the participants who provided the demonstrations. 
\bibliographystyle{IEEEtran}
\bibliography{references}

\end{document}